\documentclass{anstrans}
\title{Prediction of Critical Heat Flux in Rod Bundles Using Tube-Based Hybrid Machine Learning Models in CTF}
\author{Aidan Furlong$^{*}$, Robert Salko$^{\dagger}$, Xingang Zhao$^{\ddagger}$, Xu Wu$^{*}$}

\institute{
$^{*}$Department of Nuclear Engineering, North Carolina State University, Burlington Engineering Laboratories, 2500 Stinson Drive, Raleigh, NC 27695, ajfurlon@ncsu.edu, xwu27@ncsu.edu
\and
$^{\dagger}$Nuclear Energy and Fuel Cycle Division, Oak Ridge National Laboratory Oak Ridge, TN, USA 37830, salkork@ornl.gov
\and
$^{\ddagger}$Department of Nuclear Engineering, University of Tennessee, 823 Neyland Dr, Knoxville, TN 37996, xzhao47@utk.edu
}

\usepackage{graphicx} 
\usepackage{booktabs} 
\usepackage{microtype} 
\usepackage{threeparttable}
\usepackage[table]{xcolor}
\usepackage{siunitx}
\usepackage{upgreek}

\newcommand\blfootnote[1]{%
  \begingroup
  \begin{NoHyper}
  \renewcommand\thefootnote{}\footnote{#1}%
  \addtocounter{footnote}{-1}%
  \end{NoHyper}
  \endgroup
}

\begin{document}
\blfootnote{
This manuscript has been authored by UT-Battelle, LLC, under contract DE-AC05-00OR22725 with the US Department of Energy. The publisher acknowledges the US government license to provide public access under the DOE Public Access Plan (http://energy.gov/downloads/doe-public-access-plan).
}
\section{Introduction}

The prediction of critical heat flux (CHF) using machine learning (ML) approaches has become a highly active research activity in recent years~\cite{qi2025machine}, the goal of which is to build models more accurate than current conventional approaches such as empirical correlations or lookup tables (LUTs). Most of the recent work has considered tube-based geometries, motivated by the release of the 24,579-point US Nuclear Regulatory Commission (NRC) CHF database used to generate the 2006 Groeneveld LUT~\cite{groeneveld2019critical}. Besides purely data-driven ML methods, hybrid techniques have also been developed to use residual-predicting ML models to correct biases in low-fidelity base models (such as correlations).

Previous work~\cite{furlong2025deployment} developed and deployed tube-based pure and hybrid ML models in the CTF subchannel code~\cite{salko2023ctf} and validated them against a holdout partition of the NRC CHF database and the Bennett/Harwell test series. These models used an inlet condition--based formulation, which implicitly assumes an isolated flow channel in which heat balance assumptions are satisfied. ML-based models performed exceptionally well, with one model observed to have 2.90\% mean absolute relative error on a 2,458-point test set.

The prediction of CHF in single-channel tubes is valuable, but full-scale reactor core simulations require the use of rod bundle geometries. Unlike isolated subchannels, rod bundles experience complex thermal hydraulic phenomena such as channel crossflow, spacer grid losses, and effects from unheated conductors. Although available ML approaches may be sufficient to natively solve this problem, the limited volume of public data is prohibitive for training high-performance bundle-specific models. Before adopting methodologically complex approaches such as transfer learning, preliminary analysis is needed to determine how a local formulation tube-trained ML model behaves when predicting CHF in bundles.

This study investigates the application of both hybrid and purely data-driven ML methods, originally trained using tube-based CHF data, to predict the location and magnitude of CHF in a reference rod bundle. Three ML-based CHF models were implemented within the CTF subchannel code for evaluation: a purely data-driven ML model, a hybrid Bowring model, and a hybrid LUT model. Model performance was assessed using the Combustion Engineering (CE) 5$\times$5 test series~\cite{karoutas2004subcooled}, which is included within CTF's validation suite. The overall objective of this work is to quantify how well these otherwise \textit{unadapted} CHF models generalize to a more complex geometry and environment where the physical assumptions implicit in the tube-based training data are no longer guaranteed.

\section{Methods}

Three ML-based models were used in this study: a purely data-driven deep neural network (DNN) and two hybrid DNN-based arrangements. In the canonical hybrid approach, a prediction workflow begins with a base model (typically an empirical correlation or LUT), $f_{\mathrm{base}}(x)$, which provides a first estimate for CHF, $\hat{y}_i$. An ML model, $f_{\mathrm{ML}}(x)$, trained using the residual between $\hat{y}_i$ and the known experimental value $CHF_{\mathrm{exp},i}$ as the target, then outputs the predicted residual, $\hat{r}_i$. This value is then used to adjust the base model's low-fidelity estimate, yielding the final hybrid prediction $CHF_{\mathrm{pred},i} = \hat{y}_i + \hat{r}_i$. In this work, the Bowring correlation~\cite{bowring1972simple} and Groeneveld LUT~\cite{groeneveld20072006} were used as base models.

\subsection{Tube-Based ML Models}

All training data used in this study originate from the NRC CHF database used to generate the 2006 Groeneveld LUT~\cite{groeneveld2019critical}, which is made up entirely of tube-based upflow water experiments with uniform axial power profiles. The 24,579 entries were filtered to exclude entries with negative inlet subcooling values, leaving a final cleaned tube dataset of 24,320 points. This dataset was shuffled, partitioned into training/validation/testing datasets using a 90/05/05 split, and then standardized. This training-weighted split was chosen to maximize performance as deployment models; the dataset is also of adequate size to support evaluation with 5\%.

The three ML models use a fully local input feature set: heated equivalent diameter ($D_{\mathrm{he}}$), pressure ($P$), mass flux ($G$), and local equilibrium quality ($x_{\mathrm{e}}$). This formulation, $CHF(D_{\mathrm{he}},P,G,x_{\mathrm{e}})$, was chosen to avoid explicit/implicit dependencies on inlet conditions, as would be the case if inlet enthalpy subcooling ($\Delta h_{\mathrm{sub}}$) or heated length ($L_{\mathrm{h}}$) were included. The addition of these parameters significantly increases ML model performance in isolated subchannel environments (like tubes) but can be detrimental in settings where crossflow mixing between neighboring subchannels occurs (essentially all rod bundles), if those effects are not captured too. It should be noted that $D_{\mathrm{he}}=D_{\mathrm{hy}}=D_{\mathrm{tube}}$ in the training data; $D_{\mathrm{he}}$ is shown to indicate its use in CTF as an input feature.

The depths and widths of the three model architectures were determined independently via Bayesian hyperparameter optimization, along with other hyperparameters such as batch size, learning rate, and L2 regularization strength. The best hyperparameter configurations were then trained to a maximum of 500 epochs, employing exponential learning rate decay and early stopping to reduce the risk of overfitting. These three final models were then exported from the Python environment and loaded into CTF using an in-house, open-source framework developed by the authors in a prior work~\cite{furlong2025nativefortranimplementationtensorflowtrained}.

Validation after transfer to CTF was then performed using reference CTF models built from the 1,216-point holdout partition of the NRC CHF tube database. Each of these simulations was made up of 60 axial nodes and run as transients to steady state over 20 simulation seconds with CHF computed at the exit node. Six key statistics for these runs are reported in Table \ref{tab:comparison_nrc}: mean absolute percentage error ($\upmu_\text{error}$), median absolute percentage error ($\text{Med}_{\text{error}}$), maximum absolute percentage error ($\text{Max}_{\text{error}}$), standard deviation of the absolute percentage error ($\text{Std}_{\text{error}}$), the proportion of absolute percentage error values above 10\% ($F_{\epsilon>10\%}$), and the proportion of absolute percentage error values above 25\% ($F_{\epsilon>25\%}$).

\begin{table}[ht!]
    \centering
    \caption{Performance of final CTF tube-based ML models, evaluated on the NRC \textit{tube} database test partition.}
    \label{tab:comparison_nrc}
    \resizebox{\linewidth}{!}{
    \begin{tabular}{lccc}
        \toprule
        Metric & Pure ML & Hybrid Bowring & Hybrid LUT \\
        \midrule
        $\upmu_\text{error}$ (\%)            & 11.63 & 12.69 & 11.63 \\
        $\text{Med}_{\text{error}}$ (\%)     & 6.03 & 6.46 & 6.05 \\
        $\text{Max}_{\text{error}}$ (\%)     & 294.08 & 395.00 & 333.39 \\
        $\text{Std}_{\text{error}}$ (\%)     & 18.58 & 21.98 & 19.12 \\
        $F_{\epsilon>10\%}$ (\%)             & 34.79 & 35.53 & 34.46 \\
        $F_{\epsilon>25\%}$ (\%)             & 16.37 & 17.27 & 15.80 \\
        \bottomrule
    \end{tabular}}
\end{table}

\subsection{CE 5$\times$5 Bundle Test Series}

To test tube-based ML model performance in rod bundle geometries, the CE 5$\times$5 test series was selected. This testbed is included in the documented CTF validation cases~\cite{ctf_validation} for heat transfer and turbulent mixing models but has not been used to simulate the original report's~\cite{karoutas2004subcooled} CHF experiments. The test bundle consisted of 25 rods with a heated length of 84 in. (2.13 \si{\meter}), uniform axial power profile, and non-uniform radial power distribution with peaking on rod 25. The bundle had five grid spacers along the heated length, with four thermocouple levels beginning at an elevation of 76.63 in. (1.95 \si{\meter}).

\begin{figure}[ht!]
    \centering
    \includegraphics[width=.85\linewidth]{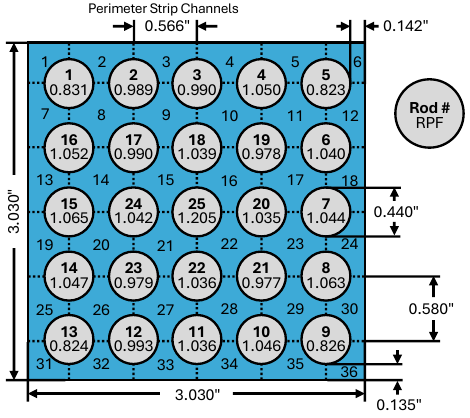}
    \caption{Lattice configuration for TS74~\cite{karoutas2004subcooled}. Note that TS75 maintains identical dimensions except for the perimeter strip channel pitch and exterior \textit{x}-axis gaps.}
    \label{fig:bundle_lattice}
\end{figure}

The experimental CHF runs consisted of eight cases between two groups, TS74 and TS75. The lattice design of the test bundle for TS74 is presented in Figure \ref{fig:bundle_lattice}. Both TS74 and TS75 had a single set of ``perimeter strip channels'' to simulate the interface of two different CE 14$\times$14 fuel assemblies. In all eight tests, CHF was detected at thermocouple level four on surface three of hot pin 25. In two cases, a temperature excursion was also observed on surface one of the same pin.

\begin{figure}[ht!]
    \centering
    \includegraphics[width=\linewidth]{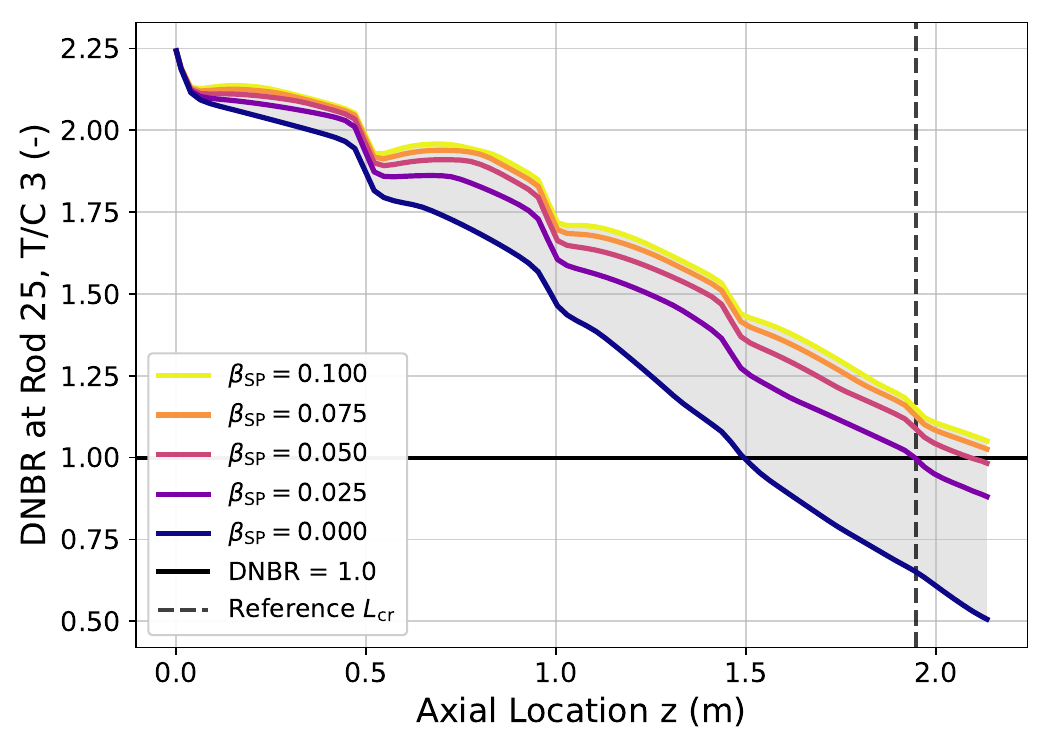}
    \caption{$\beta_{\mathrm{SP}}$ sensitivity runs for TS74.1 using the baseline LUT.}
    \label{fig:ts744c1_bLUT_beta_sensitivity}
\end{figure}

Each simulation model consisted of 25 heated rods, 36 subchannels with active crossflow, 84 axial nodes, and 5 non-mixing spacer grids. CTF's turbulent mixing model was enabled, requiring the user to specify the single-phase mixing coefficient $\beta_{\mathrm{SP}}$. Sensitivity studies have shown that this problem-dependent parameter can have significant impact on CTF CHF predictions. To illustrate this, Figure \ref{fig:ts744c1_bLUT_beta_sensitivity} shows the departure from nucleate boiling ratio (DNBR) versus heated elevation curves for various $\beta_{\mathrm{SP}}$ value choices, which can lead to differences in CHF at the experimental location in excess of 50\%. A coefficient value of $\beta_{\mathrm{SP}}$ = 0.0044 was selected, which was optimized specifically for this testbed in a previous study~\cite{sung2014application}.

Simulations were performed for all eight experimental cases, with six CTF runs in each: three ML-based CHF models and three baseline CHF models. The W-3 correlation~\cite{tong1967heat}, built specifically for rod bundles, was used as a baseline in addition to the Bowring correlation and Groeneveld LUT.

\section{Results}

After each simulation terminated successfully, the DNBR and $q''_{\mathrm{local}}$ values were collected from all pin surfaces at all elevations; CHF values were computed as the product of the two. Since these experiments have axially uniform heating, $q''_{\mathrm{local}}$ will be roughly constant over the heated length. Although CHF was observed experimentally at thermocouple 3 of rod 25 at an elevation of 1.95 \si{\meter}, there is no guarantee that the corresponding CTF subchannel will also be the limiting subchannel. Thus, for each case, the subchannels corresponding to the limiting experimental channel were verified to exhibit the minimum DNBR and therefore also be limiting. Two key quantities were then extracted: the predicted CHF location, defined as when DNBR crossed below 1.0, and the predicted CHF magnitude computed at the \textit{experimentally observed} location (1.95 \si{\meter}~above the base).

These two predicted quantities were then compared against their experimental values to measure CHF model performance and are reported in Table \ref{tab:ce_results}. In all but one test case, the runs using the hybrid LUT CHF model achieved the lowest error in both predicted quantities. Case 3 for both TS74 and TS75 led to a break in trend for all CHF model runs, with baseline models exhibiting smaller error metrics than in other tests and the baseline Bowring even outperforming all other CHF model groups in TS74.3. On average, however, CTF runs using ML-based CHF models were observed to have more accurate predictions than baseline comparators.

\begin{table}[ht!]
    \centering
    \caption{Relative error metrics for the CE 5$\times$5 runs. Upper and lower cell values respectively denote relative error (\%) of the predicted CHF \textit{value} and the predicted CHF \textit{location}.}
    \label{tab:ce_results}
    \resizebox{\linewidth}{!}{
    \begin{tabular}{lcccccc}
        \toprule
        Case & Base & Base & Base & Pure & Hybrid & Hybrid \\
             & W-3 & Bowring & LUT & ML & Bowring & LUT \\ \midrule

        TS74.1 & -25.10 & -26.68 & -23.18 & -12.19 & -15.79 & \cellcolor{gray!25}6.15 \\
                    & -20.73 & -15.94 & -18.16 & -12.22 & -9.99 & \cellcolor{gray!25}2.11 \\
        \midrule
        TS74.2 & -23.16 & -23.05 & -22.92 & -18.05 & -14.76 & \cellcolor{gray!25}4.17 \\
                    & -18.13 & -13.56 & -16.34 & -8.85 & -7.97 & \cellcolor{gray!25}1.21 \\
        \midrule
        TS74.3 & -15.87 & \cellcolor{gray!25}-5.91 & -13.51 & -12.42 & -11.75 & -14.76 \\
                    & -16.20 & \cellcolor{gray!25}-5.28 & -10.75 & -11.43 & -8.96 & -9.72 \\
        \midrule
        TS74.4 & -31.64 & -40.18 & -24.13 & -13.56 & -26.19 & \cellcolor{gray!25}13.52 \\
                    & -25.92 & -24.69 & -23.05 & -11.41 & -16.47 & \cellcolor{gray!25}6.44 \\
        \midrule
        TS75.1 & -27.00 & -27.19 & -23.70 & -16.47 & -16.83 & \cellcolor{gray!25}2.90 \\
                    & -23.71 & -16.77 & -19.73 & -16.83 & -12.38 & \cellcolor{gray!25}0.88 \\
        \midrule
        TS75.2 & -23.35 & -20.93 & -21.44 & -13.76 & -15.32 & \cellcolor{gray!25}3.19 \\
                    & -20.19 & -12.80 & -16.72 & -7.03 & -8.93 & \cellcolor{gray!25}0.94 \\
        \midrule
        TS75.3 & -16.91 & -6.40 & -13.48 & -16.32 & -5.75 & \cellcolor{gray!25}-5.71 \\
                    & -18.05 & -5.67 & -11.28 & -17.55 & -4.25 & \cellcolor{gray!25}-3.65 \\
        \midrule
        TS75.4 & -36.28 & -44.58 & -28.09 & -15.54 & -28.24 & \cellcolor{gray!25}0.91 \\
                    & -32.10 & -28.28 & -27.10 & -16.42 & -19.11 & \cellcolor{gray!25}0.66 \\
        \midrule
        \textbf{Mean} & 24.91 & 24.37 & 21.31 & 14.79 & 16.83 & \cellcolor{gray!25}1.30 \\
         & 21.88 & 15.37 & 17.89 & 12.72 & 11.01 & \cellcolor{gray!25}-0.14 \\
        \bottomrule
    \end{tabular}}
\end{table}

\begin{figure*}[ht!]
    \centering
    \includegraphics[width=0.85\linewidth]{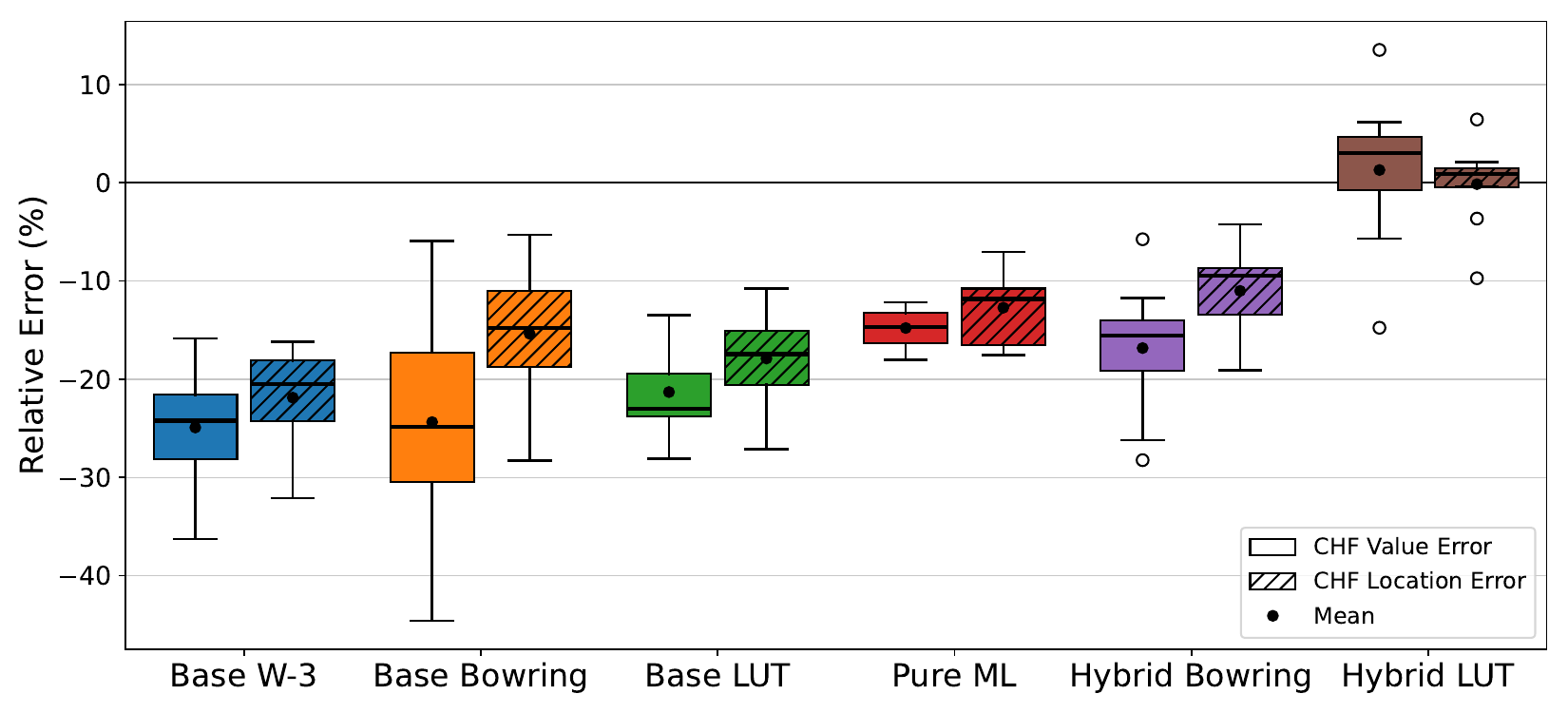}
    \caption{Boxplot comparison of relative error values between each of the CHF model groups.}
    \label{fig:boxplot_comparison}
\end{figure*}

To better visualize the distributions of the models' error values, boxplots are presented in Figure \ref{fig:boxplot_comparison}, color-coded for CHF model type and patterned to indicate whether the error corresponds to CHF magnitude or location. All three baseline models maintain similar mean and median values for CHF magnitude (between 21\% and 25\%), but the baseline Bowring exhibits a significantly larger interquartile range and whiskers. Of the baseline models, the cases using the Bowring base achieved the smallest mean and median relative error in terms of CHF location predictions, a value of 15.37\%. Reduced mean and median relative errors were observed for each ML-based group in terms of both magnitude and location.

\begin{figure}[ht!]
    \centering
    \includegraphics[width=\linewidth]{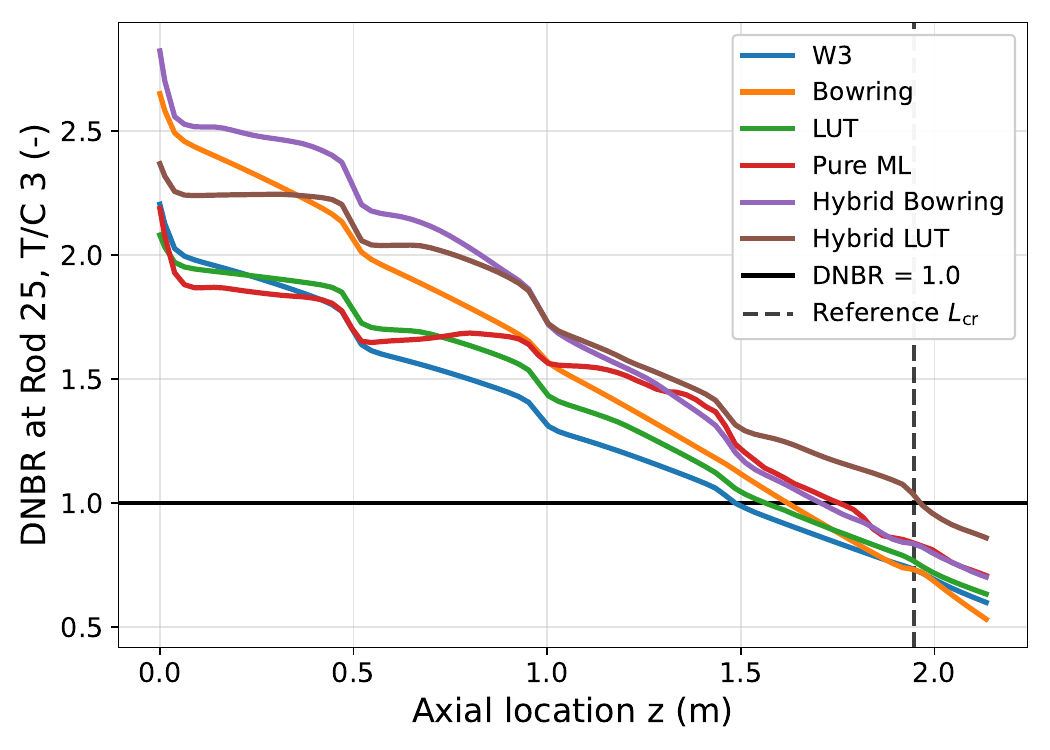}
    \caption{Axial DNBR curves for TS75 Case 1 comparing CHF modeling approaches.}
    \label{fig:ts74c1_axial_dnbr}
\end{figure}

Although these error metrics are valuable as single points of comparison between models, it is also necessary to assess the \textit{qualitative} behavior of the CHF predictions being made. To do so, the hot channel DNBR curves for each of the model types were plotted with respect to axial location, with TS75 Case 1 presented in Figure \ref{fig:ts74c1_axial_dnbr}. The dashed line ``Reference $L_{\mathrm{cr}}$'' is the elevation of the thermocouple at which CHF was experimentally detected, 76.63 in. (1.95 \si{\meter}).

All curves exhibit decreases in DNBR over the length of the bundle; sharp decreases occur at spacer locations, which are coincident with local pressure and mass flux drops. The spread of initial DNBR values is large, between 2.1 and 2.8. All three baseline cases fall in DNBR with relatively consistent overall slopes, baseline Bowring having the largest. The pure ML curve is less predictable, with a changing moving average, especially between 0.5 and 1.5~m. The hybrid Bowring curve, in general, parallels its baseline counterpart but remains larger. The hybrid LUT trajectory is similar in shape to that of the baseline LUT but remains about 0.25 larger in DNBR and nearly intercepts the experimental critical location.

Of all CHF model groups, the hybrid LUT produced predictions with consistently smaller error for both CHF magnitude and location. The axial behavior of the hybrid LUT's predictions was similar to the base model's, but the hybrid's ML component corrected its magnitude to better match that observed experimentally. It should be noted that in six of the eight cases the hybrid LUT was positively biased, an undesired trait when considering conservatism in a practical setting.

\section{Conclusions}

This study investigated the generalization of ML-based CHF prediction models in rod bundles after being trained on tube-based CHF data. A purely data-driven DNN and two hybrid bias-correction models were implemented in the CTF subchannel code and used to predict CHF location and magnitude in the CE 5$\times$5 CHF test series. The W-3 correlation, Bowring correlation, and Groeneveld LUT were used as baseline comparators. On average, all three ML-based approaches produced magnitude and location predictions more accurate than the baseline models. Between cases, however, baseline models achieved smaller errors than ML-based counterparts in some instances. Overall, the hybrid LUT exhibited the highest accuracy, albeit with a slight positive bias in magnitude predictions: a mean relative error of 1.30\% in CHF magnitude and $-$0.14\% in location.

While these results are encouraging, it is important to note that subchannel simulations contain many sources of uncertainty that have the potential to affect the observed performance of individual modeling components, including CHF models. Future work will primarily focus on acquiring and processing additional rod bundle data, which remains a significant limitation. An increase in data availability would allow for a more structured feature engineering process and the application of traditional analysis and optimization techniques not feasible under severe data scarcity. Outside additional data, transfer learning approaches offer a principled framework for leveraging the existing tube and annulus data while also adapting models to bundle-specific phenomena.

\section{Acknowledgments}

\noindent The authors from the North Carolina State University were also funded by the U.S. DOE Office of Nuclear Energy Distinguished Early Career Program (DECP) under award number DE-NE0009467.

\bibliographystyle{ans}
\bibliography{bibliography}
\end{document}